# Automatic detection of aerial survey ground control points based on Yolov5-OBB

CHENG Chuanxiang[1], YANG Jia [1]*, WANG Chao[2], ZHENG Zhi[3], LI Xiaopeng[4], DONG Di[5], CHANG Mengxia[1], ZHUANG Zhiheng[6],

*Abstract*—The use of ground control points (GCPs) for georeferencing is the most common strategy in unmanned aerial vehicle (UAV) photogrammetry, but at the same time their collection represents the most time-consuming and expensive part of UAV campaigns. Recently, deep learning has been rapidly developed in the field of small object detection. In this letter, to automatically extract coordinates information of ground control points (GCPs) by detecting GCP-markers in UAV images, we propose a solution that uses a deep learning-based architecture, YOLOv5-OBB, combined with a confidence threshold filtering algorithm and an optimal ranking algorithm. We applied our proposed method to a dataset collected by DJI Phantom 4 Pro drone and obtained good detection performance with the mean Average Precision (AP) of 0.832 and the highest AP of 0.982 for the cross-type GCP-markers. The proposed method can be a promising tool for future implementation of the end-to-end aerial triangulation process.

*Key Words*—UAV tilt photogrammetry, automatic detection of GCPs, deep learning, YOLOv5-OBB, aerial triangulation

## I. INTRODUCTION

Unlike conventional surveys involving huge costs, labor, and time, unmanned aerial vehicles (UAVs) photogrammetry is a cost-effective way to conduct aerial surveys at ultra-high spatial resolutions (1 cm to 1 m) for numerous applications in the close-range domain [1], [2]. To ensure the geometric accuracy of derived maps and other data products in a map coordinate system, it is essential to use accurate location data to align or georeference the captured UAV imagery. In general, market-available civil UAVs carry a low-cost GNSS/IMU (Global Navigation Satellite System/Inertial Measurement Unit) module and collect the position and orientation information [3].

Although it is convenient to use direct orientation for georeferencing, this method does not meet the requirements for high-precision (e.g. centimeter-level) mapping in some applications [4]–[6]. Alternative practical way (also called in-direct orientation) is to use Ground Control Points (GCPs, with known highly precise and accurate coordinates and elevation) as a "hook" to tie the captured images down to the earth's surface[7].For instance, C. Hugenholtz et al [5] showed that georeferencing with and without GCPs showed similar accuracy in the horizontal direction, but the error between the two in the vertical direction differed by a factor of 2-3. For the in-direct orientation approach, widely used GCP-markers that look distinctly different from the background are manually placed around the area of interest prior to conducting surveys [3]. The high accuracy locations (i.e. GCPs information) of GCP-markers are manually measured using a GNSS device. While obtained in exactly the same way as GCPs, the Check-points serve to verify the accuracy of the georeferenced map by comparing the GNSS-measured locations to the coordinates of the Check-points shown on the map.

Gianfranco F et al.[6] found that adding only one GCP in GCP-free UAV bundle adjustment resulted in a georeferenced image to be as good as the generated image georeferenced with only GCP. Yang, J [8] showed that the loss of GCP during the aerial survey significantly impacted the overall aerial survey results. With the development of direct orientation techniques, although we may further reduce the number of GCPs, the number of Check-points cannot be reduced when evaluating the geometric accuracy of the acquired image products [9],[10].

In a word, acquiring GCPs remains a part of inexpensive civil UAV mapping because GCPs can significantly improve the accuracy of aerial triangulation results and ensure the quality of the generated maps and other products. The general process of acquiring GCPs includes deploying GCP-markers, measuring their ground positioning information, and identifying their corresponding positions in the UAV imagery. However, GCP-markers are tiny targets on the UAV imagery, so finding GCP-markers and manually acquiring the GCPs information is a highly tedious and time-consuming task.

The whole process of UAV aerial triangulation nowadays is well addressed by an open-source algorithm, except for the GCPs automatically added to the bundle adjustment [2]. Jain et al. [11] proposed a pipeline to segment white L-shape GCP-markers from the image automatically by integrating three components of edge oriented histogram, canny edge detection, and Convolutional Neural Network (CNN) classification. However, the sophisticated processing pipeline they proposed

1. School of Surveying and Urban Spatial Information, Henan University of Urban Construction, Pingdingshan,467036,
china (email: 201912221@hncj.edu.cn;202012022@hncj.edu.cn;).
2.Department of Geological Sciences, University of North Carolina, Chapel Hill, NC, USA(e-mail: chao.wang@unc.edu)
3. the Northeast Institute of Geography and Agricultural Ecology, Chinese Academy of Sciences, Changchun, China (e-mail: zhengzhi@iga.ac.cn).
4.School of Geoscience and Technology, Zhengzhou University, Zhengzhou 450001, China(e-mail: lixp@gs.zzu.edu.cn).
5. South China Sea Institute of Planning and Environmental Research, SOA, Remote Sensing Lab; Key laboratory of Marine Environmental Survey Technology and Application(e-mail: dongdide90@163.com)
6.the Department of Surveying Engineering, Heilongjiang Institute of Technology, Harbin 150050, China(e-mail: zhz19971223@126.com).
Corresponding author: YANG Jia are with School of Surveying and Urban Spatial Information, Henan University of Urban Construction, Pingdingshan,467036, China (email: yangjia@hncj.edu.cn;).

makes it difficult to replicate. In addition, the edge oriented histogram and edge detection algorithm can easily fail when there are other objects of similar color and shape on the ground. Ren [12] used the covariance equation to locate the position of GCPs. However, they can only estimate approximate locations of GCPs because systematic errors are difficult to eliminate. It may also require some manual corrections to accurately locate each GCPs.

In recent years, deep learning techniques have been widely used and achieve state-of-the-art performance in many fields [13]. In this study, we propose an easy-to-implement workflow that integrates YOLOv5-OBB, one of compound-scaled arbitrary orientation object detection deep learning architecture, with confidence threshold filtering and optimal ranking to automatically detect GCP-markers and locate the position of GCP.

## II. METHOD

### A. GCPs dataset and data preprocessing

To explore the influence of the shape and pointing of the GCP-markers on model training, we conducted experiments with L-shape and cross-shape GCP-markers (Fig. 1). According to different locations of GCP on the GCP-marker, we considered four kinds of the L-shape GCP-markers, including top-left, bottom-left, top-right, and bottom-right. The cross-shape GCP-marker has a relatively simple GCP location (i.e., at its center) and is therefore considered as a category only. In this study, we collected about 5000 UAV images using the DJI Phantom 4 Pro drone.

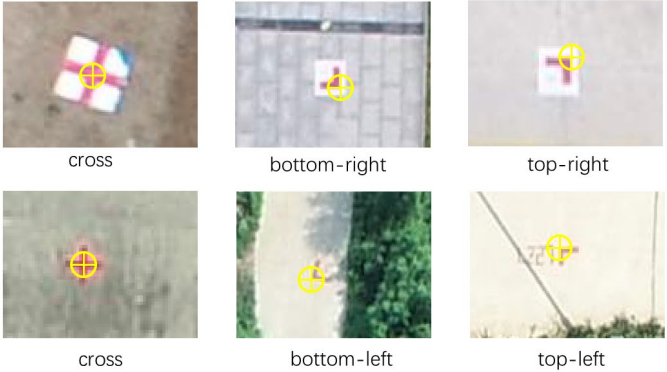

**Fig. 1.** Types of GCP markers. The yellow circle marker is the location of the GCP

In theory, the larger size of input imagery should provide more information and yield more accurate results. However, the size of the acquired UAV image is 5,472 × 3,648 pixels, which is too large to directly fit into GPU memory. It also requires a longer training time, and the added information may be redundant, which then affects the model training process as well as the subsequent detection performance. Therefore, we followed the method of YOLT [14] and cropped the original large-size images to a suitable size. To do so, we first padded the original image along both axes and then cropped it uniformly to prevent distortion. We finally acquired 2,358 cropped images with GCP-markers as the training dataset.

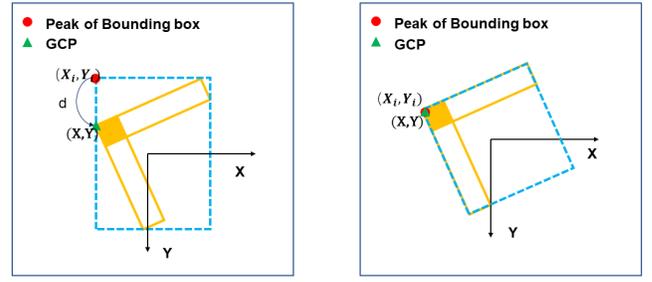

**Fig. 2.** (a) Horizontal Bounding Box. (b) Orientation Bounding Box. The yellow dot is the position of the stabbing point and the green dot is the position of the nearest vertex of the target detection frame to the stabbing point position.

### B. Detect GCP-Markers and locate the position of GCP

**1) The vertices position of arbitrary Orientation Bounding Box instead of GCP's position**

In general, when using object detection algorithms to detect ground targets, we can use four vertex coordinates of the horizontal bounding box as the GCP positions if the GCP markers can be considered axis-aligned objects. However, in many cases, objects in the UAV image are not exactly aligned to the image axis, so the resulting four vertex coordinates cannot be used as the GCP position with high precision. For example, **Fig. 2**a shows the case where the axis-aligned horizontal bounding box of this GCP-marker is not aligned with its edges, so the vertex coordinates cannot be used as GCPs. To address this issue, we adopt the arbitrary orientation object detection algorithm to generate the oriented bounding box (e.g. Fig. 2b) that better matches the outlines of the GCP-marker. Thus, the vertex position of the detected oriented bounding box in the image coordinates (x,y) matches the position of GCP and thus can be used as a proxy to the position of GCP.

YOLO-v5 (the fifth version of You Only Look Once)[15] has achieved state-of-the-art performance in small target detection. In this study, the YOLOv5-OBB[16] algorithm is an improvement on YOLO-v5 based on the circular smooth label structure[17], which can detect targets with arbitrary directions on remote sensing images. We use four vertex coordinates to represent the Orientation Bounding Box positions, where (x1,y1) is the starting point and the other points are labeled as (x2,y2), (x3,y3), and (x4,y4) in a clockwise direction in order with it as the starting point. We find that the Orientation Bounding Box used to display the detected GCP-markers starts mainly at the bottom-right, so we can use Equation (1) to calculate the position of the GCP.

$$\begin{cases} [x,y]^T = [(j-1)*w, (i-1)*h]^T + [x_1, y_1]^T, right\ down \\ [x,y]^T = [(j-1)*w, (i-1)*h]^T + [x_2, y_2]^T, left\ down \\ [x,y]^T = [(j-1)*w, (i-1)*h]^T + [x_3, y_3]^T, left\ up \quad (1) \\ [x,y]^T = [(j-1)*w, (i-1)*h]^T + [x_4, y_4]^T, right\ up \\ [x,y]^T = [(j-1)*w, (i-1)*h]^T + [\Sigma x_i, \Sigma y_i]^T/4, cross \end{cases}$$

where w, h, are the width and height of the detected image; i, j, are the number of rows of the cropped image in the original image; BR, BL, TR, TL, and CR refer to bottom-right, bottom-left, top-right, top-left and cross-shaped GCP-markers types, respectively.

**2) Detection performance in different scale GCP-markers**

During the acquisition of UAV images, the scale of GCP-markers in the images changes with the surface topography. To cope with this issue, the constructed YOLOv5-OBB model should be able to accurately identify GCP-markers independent of their scale variation. To this end, we acquired UAV images of GCP-markers of the same size, type, and color deployed on different surfaces from a relative altitude range of 60-220 meters. In this way, we can identify GCP-markers on the acquired images and evaluate the effectiveness of the established model.

*3) Confidence threshold filtering algorithm*

The complex background information in the UAV image scene is the main factor affecting the model detection performance. Artificial features, such as domestic waste, are very similar to GCP-markers, which will directly affect the accuracy of model detection. Although these false-positive features are often detected incorrectly by the trained model, their confidence values are usually small. Setting a confidence threshold can effectively filter out most of these false positive detections, but will inevitably result in some images with GCP-markers being lost. Fortunately, UAVs generally acquire images using a redundant overlay strategy to ensure the quality of subsequent image stitching, so the same GCP marker may be captured by multiple photos.

To estimate the best confidence thresholds, we selected four different sites and used GCP-markers of the same size, material, and shape with a DJI Phantom 4 Pro UAV. Also, Site 4 datasets are the images collected at different aerial heights to determine the confidence threshold statistics.

TABLE I
DATA SET FOR DETERMINING THE SIZE OF THE CONFIDENCE THRESHOLD

| | Number Of GCPS | Original Image | Cropped Image | Image Of Containing Image |
|---|---|---|---|---|
| Site 1 | 4 | 533 | 25400 | 280 |
| Site 2 | 7 | 362 | 17376 | 238 |
| Site 3 | 3 | 591 | 28368 | 281 |
| Site 4 | 6 | 259 | 15390 | 347 |

This experimental study uses a dataset of UAV images collected from four sites (Table I), acquiring a total of 1,745 images with a pixel size of 5,472 × 3,648. During training, the YOLO model generally uses input images with a pixel size of a multiple of 32. In this experiment, we cropped each image to 608 x 608 pixels tiles to allow for an easy trade-off between speed and accuracy, resulting in 86,574 images for model training and validation. In the model validation, we found that the cropped image dataset has a large number of images without GCP-markers, and the false positive GCPs tend to have low confidence level. For this reason, we can use the confidence threshold filtering method to exclude possible false positive GCPs while ensuring that images with GCP-markers are not lost. This involves the determination of the optimal confidence threshold. For this purpose, we derive equations (2) and (3). Equation (2) measures the percentage of all images in the dataset above the confidence threshold that contains GCP-markers. Equation (3) calculates the percentage of the number of lost GCPs to all GCPs due to setting the confidence threshold. True positive (TP) is the number of positive samples that are correctly predicted

$$Precision = \frac{TP}{TP + FP} \quad (2)$$

$$Loss\ Ratio = \frac{loss\ GCP\ by\ filtering}{all\ numbers\ of\ GCP} \quad (3)$$

*4) Optimal Ranking Algorithm*

UAVs carry mostly non-metric consumer-grade cameras due to their limited payload, and it is accompanied by a large amount of image distortion. UAV images are mainly affected by radial distortion, which arises because the arc of light away from the center of the lens is larger than that near the center. The closer the GCP-markers are to the edge of the image, the lower the detection accuracy due to image distortion.

$$d = \sqrt{(x_i - w/2)^2 + (y_i - h/2)^2} \quad (4)$$

$$d_{max} = \sqrt{\left(\frac{w}{2}\right)^2 + \left(\frac{h}{2}\right)^2} \quad (5)$$

$$score = \partial * \left(1 - \frac{d}{d_{max}}\right) + confidence \quad (6)$$

$$PONA = \frac{(1 - d/d_{max}) > 0.5}{(1 - d/d_{max})} \quad (7)$$

To acquire the accurately identified GCP-markers and ensure that they are located in regions with less image distortion, we use Equations (4)-(6) to accomplish this goal. Specifically, Equation (4) is used to calculate the distance from the GCP to the center point of each image. We also calculate the farthest distance from each GCP to the center of the image using Equation (5). We adopt Equation (6) to consider the effects of both image distortion and object detection confidence in a weighted manner. The $(1 - d/d_{max})$ in Equation (6) refers to the degree of image distortion at each location of GCP-markers. The more severe the image distortion is, the smaller the value. In addition, the second component of Equation (6) refers to the confidence of the identified object as GCP-marker. The higher the detection confidence, the lower the probability of a false positive detection and the higher the probability that this detected object is a GCP-marker. The $\sigma$ in Equation (6) refers to the adjustment weights used to adjust the two above mentioned numbers. In a word, the higher the weighted score value, the lower the distortion of this GCP in the image and the higher the probability of correct detection by our proposed method. It is worth noting that multiple images may contain the same GCP-marker due to the redundancy of the UAV image acquisition strategy. To acquire the highest quality of each GCP, we calculate each score value from all images containing the same GCP-marker, then rank them from largest to smallest and select the top few large scores as the true values of GCP positions to be added to the subsequent bundle adjustment. In addition, Equation (7) is used to verify the performance of the optimal ranking algorithm, if (1-d/dmax) is greater than 0.5, the GCP is considered to be located in the region of severe image distortion. The PONA in Equation (7) refers to the percentage of GCP in the image distortion area.

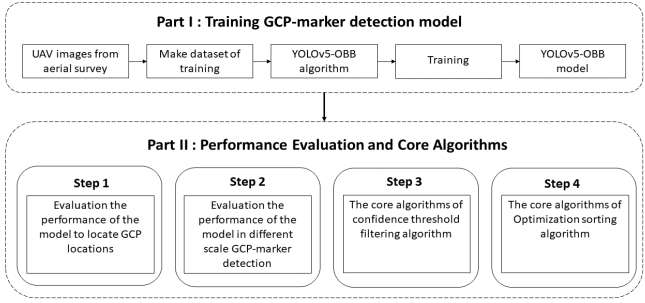

**Fig. 3.** Workflow in this study

*C. Workflow in this study*

**Fig. 3** shows the workflow of this study, which can be roughly divided into two parts. The first part focuses on preparing the GCP-markers dataset and training the YOLOv5-OBB model for detecting GCP-markers. The second part is the core of this study, which includes the evaluation of the detection performance of the model, the filtering of the detection results and the effect of optimization ranking.

1. The accuracy of the detected GCP position by YOLOv5-OBB will directly affect the accuracy of the solution of the subsequent aerial triangulation. For this reason, Step 1 compares the differences between the GCP positions identified by YOLOv5-OBB and those real GCP locations.

2. The scale of GCP-markers on the image varies with the flight altitude of the UAVs. Step 2 explores the effectiveness of the YOLOv5-OBB model in detecting GCP-markers at different scales and explores the smallest GCP-markers that can be detected.

3. The complex backgrounds in the aerial survey scene seriously affect the identification accuracy of the deep learning model. The false positive detection results can be effectively filtered by exploring the optimal confidence threshold (i.eStep 3).

4. Since the GCP-markers laid on the ground are generally acquired by multiple UAV images, they are distributed in different locations of the images. When the GCP-markers are closer to the edge of the image, adding detected GCPs to the bundle adjustment introduces errors due to the distortion of the image. To filter the GCP-markers located in the non-image distortion region, we propose the optimal ranking algorithm (i.eStep 4) and explore its application performance.

### III. RESULTS AND ANALYSIS

*A. Model performance*

Considering the tradeoff of model weight parameters, we choose the YOLOv5-OBB model with medium rather than small and large-sized initial parameters to train the GCP-markers dataset on. We used the pre-trained weights provided in [18] as initial values to converge the model to the desired level with a total of 300 epochs. **Fig. 4.** shows the results of our tests on the trained model, where the mAP (i.e. mean Average Precision) is 0.832, the highest AP (i.e. Average Precision) is 0.982 for the cross-type GCPs, and the lowest AP is 0.676 for the bottom-right category.

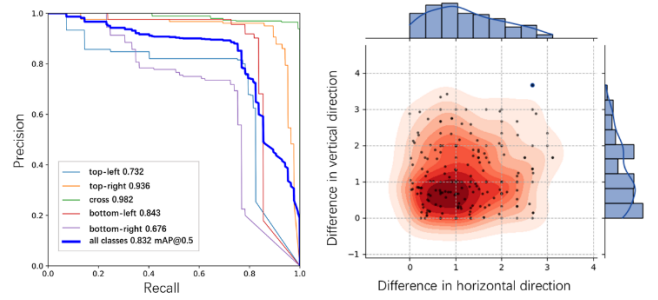

**Fig. 4. (a)** mAP of the model **(b)** the difference distribution in horizontal and vertical directions

*B. The difference between the predicted GCP position and the real position*

To quantify the accuracy of the predicted GCP locations, we plotted and analyzed error scatter (Fig. 4(b)) by randomly sampling 60% of the test datasets. Specifically, we evaluated the difference between the real GCP position and the GCP position detected by YOLOv5-OBB. Our analysis found that the maximum error of the predicted GCP positions by YOLOv5-OBB was no more than 4 pixels compared with the real GCP positions, with 80% of them within 2 pixels and 98% within 3 pixels, and only a few had relatively large errors. Our proposed method achieved a state-of-the-art accuracy level, i.e. comparable to the 1 to 3 pixel error by manually extracted positions of GCP. In addition, we found that the larger error comes from the L-type GCP marker because it has four categories and is easily confused in the YOLOv5-OBB model. In contrast to the L-type GCP marker, we recommend using the cross-type GCP marker in aerial surveys because it has only one category.

TABLE II
DETECTION PERFORMANCE OF MODELS IN DIFFERENT FLIGHT HEIGHT RANGES

| Flight altitude | GCP-markers | True Positive (TP) | False Positive (FP) | False Negative (FN) | Precision | Recall | Average GCP-markers pixel |
|---|---|---|---|---|---|---|---|
| 40-60 | 19 | 12 | 3 | 7 | 0.8 | 0.63 | 59*62.5 |
| 60-80 | 31 | 29 | 0 | 2 | 1 | 0.93 | 40.5*45.5 |
| 80-100 | 39 | 39 | 1 | 0 | 0.975 | 1 | 31.5*45.5 |
| 100-120 | 38 | 38 | 1 | 0 | 0.974 | 1 | 32.5*34 |
| 120-140 | 45 | 45 | 0 | 0 | 1 | 1 | 26*28.5 |
| 140-160 | 96 | 70 | 0 | 26 | 1 | 0.72 | 20*23 |
| 160-180 | 58 | 53 | 0 | 5 | 1 | 0.91 | 19*21 |
| 180-200 | 55 | 54 | 0 | 1 | 1 | 0.98 | 17*19 |
| 200-220 | 6 | 5 | 0 | 1 | 1 | 0.83 | 14*16 |

*C. Testing of different scale GCP-markers in model*

To reduce the confusion caused by other factors, we filtered the dataset with GCP-markers of different altitude ranges and analyzed the model performance only on these filtered data. Our results show (TABLE II) the detection performance of GCP-markers at different relative height values for eight 20 m intervals between 60 and 220 m. The overall results show that the YOLOv5-OBB model has a high performance in detecting GCP-markers at different scales, with most of them having a precision greater than 97.4% and a recall greater than 91%. Exceptionally, a low recall (of 72%) occurs in the relative altitude range of 140-160 m, corresponding to 26 images in which GCP-markers are not detected. This is because our training dataset is mostly collected from relatively flat landscapes (i.e. roads and settlements), while most of the incorrectly detected GCP-markers in this relative altitude range (i.e. 140-160m) are collected from sloping landscapes (i.e., hillside). It should be noted that the training dataset including images of different landscapes will improve the robustness of the model. In addition, we found that the smallest GCP-marker size detected by the YOLOv5-OBB model at 60-220 m relative altitude is greater than 12 × 12 pixels.

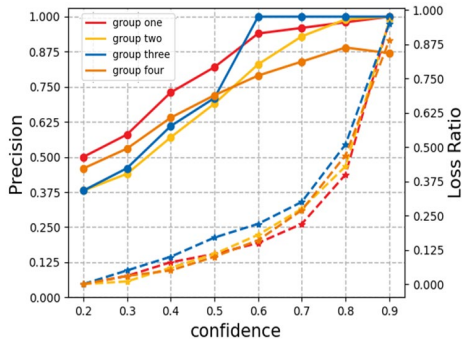

**Fig. 5.** Variation of loss-ratio and precision under different confidence thresholds

*D. Performance of confidence threshold filtering algorithm*

The YOLOv5-OBB model gives a confidence value (i.e. 0-1) when detecting GCP-markers. We can find an optimal confidence threshold to filter false-positive detections. To do so, we can perform a sensitivity analysis by looking at the trade-off between the Precision and Loss-Ratio values while varying the confidence value. Our analysis (Fig. 5) shows that as the confidence threshold increases, the Precision (i.e solid line) increases at all four sites, and the Loss Ratio (i.e., dashed line) also increases.

The confidence filtering analysis (**Fig. 5**) shows that when the confidence threshold is set to 0.7, the average Precision is relatively high, while the Loss Ratio is not too high. Only about 26% of the discarded data is tagged with GCP-markers. Statistics of the discarded data with GCP-markers reveal that 60% are data with severe image distortion. Our analysis also shows that the number of detected GCP-markers on the ground does not decrease after filtering, implying that the discarded data itself is redundant.

TABLE III
VALUES OF PRECISION AND LOSS RATIO FOR DIFFERENT GROUPS AT A CONFIDENCE THRESHOLD OF 0.7

|  | Site 1 | Site 2 | Site 3 | Site 4 |
|---|---|---|---|---|
| **Precision** | 0.93 | 0.96 | 1 | 0.84 |
| **Loss Ratio** | 0.275 | 0.22 | 0.3 | 0.27 |
| **Number of GCPs** | 203 | 185 | 199 | 251 |

Table III shows the Loss Ratio values, the Precision values, and the number of images with GCP-markers for the four data sets when the confidence threshold is set to 0.7. Even if 26% of the data with GCP-markers are discarded, the UAV images with the same GCP-markers still have a lot of redundant data, so here we do not discuss the recall metric of the model.

TABLE IV
OPTIMAL SORTING ALGORITHM PERFORMANCE

| Number | Top3 precision | Top5 precision | Top5 PONA | Top8 precision | Top10 precision | Top10 PONA |
|---|---|---|---|---|---|---|
| GCP1 | 33 | 100% | 100% | 100% | 100% | 100% | 100% |
| GCP2 | 71 | 100% | 100% | 100% | 100% | 100% | 100% |
| GCP3 | 50 | 100% | 100% | 100% | 100% | 90% | 100% |
| GCP4 | 49 | 100% | 100% | 100% | 100% | 100% | 100% |
| GCP5 | 24 | 100% | 100% | 100% | 100% | 100% | 100% |
| GCP6 | 19 | 100% | 100% | 100% | 100% | 100% | 70% |
| GCP7 | 28 | 100% | 100% | 100% | 100% | 90% | 90% |
| GCP8 | 22 | 100% | 100% | 100% | 100% | 100% | 90% |
| GCP9 | 36 | 100% | 100% | 100% | 100% | 100% | 100% |
| GCP10 | 21 | 100% | 100% | 100% | 100% | 100% | 70% |
| GCP11 | 26 | 100% | 100% | 100% | 100% | 100% | 90% |

*E. Performance of the optimal ranking algorithm*

We use the optimal ranking algorithm to filter out redundant data located in the image distortion area. We randomly selected 11 GCP markers placed on the ground in Site 1-4 as testing data to evaluate its performance.

To find the GCP with the smallest distortion region of the image, we first set the $\partial$ in equation (6) to 2. We then calculate all the scored values with the same GCP-marker in the data and rank these calculated scores from largest to smallest. In the top three values of the scores (Table IV), the accuracy of the model detection is 100% and the top five GCPs are all in the non-distorted region of the image. However, in the top 10 values within the highest scores, the model appears to have some detection errors, because there are some cases where GCPs are in the distorted regions of the images. Therefore, we recommend selecting the top 5 highest scoring GCPs to be added to the bundle adjustment.

The results show that the optimal ranking algorithm can find both the correct GCP-markers and the GCPs in the lighter image distortion region.

IV. CONCLUSION

In this letter, we use the YOLOv5-OBB combined with a confidence threshold filtering algorithm and optimal ranking algorithm to automatically detect GCP-marks and find GCP positions. The aim is to reduce the manual workload during aerial triangulation processing and to improve the efficiency of building 3D models.